\documentclass[sigconf, preprint]{acmart}
\AtBeginDocument{%
  }

\setcopyright{rightsretained}
\copyrightyear{2026}
\acmYear{2026}
\acmConference[Technical Report]{Research done at IBM. Published}{Published Feb 2026}{Yorktown Heights, NY}
\acmBooktitle{Technical Report describing research done at IBM}
\acmISBN{}

\usepackage{tikz}
\usetikzlibrary{arrows.meta, positioning, calc, shapes.geometric, fit, backgrounds}
\usepackage{listings}
\begin{document}

\title{Trajectory-Informed Memory Generation for Self-Improving Agent Systems}

\author{Gaodan Fang, Vatche Isahagian, K. R. Jayaram, Ritesh Kumar, Vinod Muthusamy, Punleuk Oum, Gegi Thomas}
\authornote{Author names listed alphabetically.}
\affiliation{%
  \institution{Agents and Automation Lab, IBM Research}
  \country{USA}}

\renewcommand{\shortauthors}{Fang et al.}

\begin{abstract}
LLM-powered agents face a persistent challenge: learning from their execution experiences to improve future performance. While agents can successfully complete many tasks, they often repeat inefficient patterns, fail to recover from similar errors, and miss opportunities to apply successful strategies from past executions. We present a novel framework for automatically extracting actionable learnings from agent execution trajectories and utilizing them to improve future performance through contextual memory retrieval. Our approach comprises four components: (1) a Trajectory Intelligence Extractor that performs semantic analysis of agent reasoning patterns, (2) a Decision Attribution Analyzer that identifies which decisions and reasoning steps led to failures, recoveries, or inefficiencies, (3) a Contextual Learning Generator that produces three types of guidance—strategy tips from successful patterns, recovery tips from failure handling, and optimization tips from inefficient but successful executions—and (4) an Adaptive Memory Retrieval System that injects relevant learnings into agent prompts based on multi-dimensional similarity. Unlike existing memory systems that store generic conversational facts, our framework understands execution patterns, extracts structured learnings with provenance, and retrieves guidance tailored to specific task contexts. Evaluation on the AppWorld benchmark demonstrates consistent improvements, with up to 14.3 percentage point gains in scenario goal completion on held-out tasks and particularly strong benefits on complex tasks (28.5~pp scenario goal improvement, a 149\% relative increase).
\end{abstract}




\maketitle

\section{Introduction}

Large Language Model (LLM) powered agents have enabled increasingly sophisticated automation of tasks ranging from web navigation to API orchestration. These agents operate by iteratively reasoning about tasks, selecting actions, executing them, and observing results. However, a fundamental limitation persists: \emph{Agents have amnesia} because most LLMs are stateless. Agents lack systematic mechanisms to learn from their execution experiences~\cite{zhang2024survey, du2025rethinking}. An agent that struggles with a particular API authentication flow today will struggle with the same flow tomorrow unless its prompts are manually updated. An agent that discovers an efficient strategy for a task cannot automatically apply that strategy to similar future tasks. An agent that successfully recovers from an error provides no benefit to future executions that encounter similar errors.

Consider a simple e-commerce task: adding items to a shopping cart and completing checkout. An agent might successfully complete this task but do so inefficiently—for instance, by calling \texttt{amazon\_remove\_from\_cart(item\_id)} in a loop to empty the cart when a single \texttt{amazon\_empty\_cart()} call would suffice. In another execution, the agent might fail entirely because it attempts checkout without first adding a payment method, then successfully recover by recognizing the error and adding payment information. In yet another execution, the agent might execute the task cleanly from the start by systematically verifying prerequisites before each operation.

Each of these trajectories contains valuable learnings (for future executions), but of different types. The inefficient success suggests an \textbf{optimization tip}: when emptying a cart with multiple items, use the bulk operation rather than iterating through individual removals. The failure-then-recovery suggests a \textbf{recovery tip}: when checkout fails due to missing payment method, verify payment information is configured before retrying. The clean success suggests a \textbf{strategy tip}: before initiating checkout operations, systematically verify all prerequisites including cart contents, shipping address, and payment method availability.

Current approaches to agent improvement are inadequate for capturing these diverse learning opportunities. Rule-based systems require developers to manually anticipate patterns and encode them as decision rules, making them brittle and unable to adapt to unforeseen situations. Prompt engineering improves common patterns through iteratively refined instructions and examples, but this guidance is generic rather than specific to actual deployment experiences, and there is no mechanism for automatic improvement based on observed outcomes. Generic memory systems~\cite{chhikara2025mem0, xu2025amem} store facts extracted from conversations in vector databases for later retrieval, but these systems lack several critical capabilities: they have no understanding of agent execution patterns and reasoning flows, they cannot perform causal analysis to identify which decisions led to failures or inefficiencies, they lack structured learning extraction with categories like strategy, recovery, and optimization, and they provide no provenance tracking from learnings back to source trajectories. Recent work has begun extracting reusable knowledge from agent trajectories---including workflows from successful executions~\cite{wang2024awm, feng2025agentrr}, procedural instructions~\cite{fang2025memp}, reasoning strategies~\cite{cai2025reasoningbank}, and evolving context playbooks~\cite{zhang2025ace}---but these approaches typically learn only from successful trajectories, lack explicit causal attribution of failures, or produce monolithic documents rather than structured, retrievable memory entries. Empirical studies further demonstrate that naive experience accumulation leads to error propagation and misaligned replay~\cite{xiong2025memory}, underscoring the need for quality-aware memory curation.

We present a framework that addresses these limitations through trajectory-informed memory generation and retrieval. Our key insight is that agent execution histories—trajectories—contain rich semantic information about not just what happened, but why agents made decisions, how they reasoned about tasks, which strategies succeeded, which patterns proved inefficient, and where decision chains led to failures and recoveries. By analyzing these trajectories with semantic understanding, we can automatically extract actionable learnings across multiple categories, attribute failures and inefficiencies to specific decisions and reasoning steps, generate context-aware guidance, and retrieve relevant learnings based on multiple contextual dimensions.

Our contributions are as follows:

\begin{itemize}
\item We introduce trajectory intelligence extraction that moves beyond raw logging to semantic understanding of agent reasoning patterns, including analytical thoughts, planning patterns, validation behaviors, reflection patterns, and self-correction sequences.

\item We present automated decision attribution that distinguishes immediate causes, proximate causes, and root causes of failures, while also identifying which decisions led to successful recoveries and which execution patterns prove inefficient despite succeeding.

\item We develop contextual learning generation that produces three distinct types of guidance: strategy tips encoding successful patterns from clean executions, recovery tips capturing failure handling and error correction approaches, and optimization tips identifying efficiency improvements from successful but suboptimal executions.

\item We design adaptive memory retrieval that combines semantic similarity with metadata filtering and priority-based ranking to ensure agents receive the most relevant guidance for their specific context, including task type, domain, and execution patterns.

\item We demonstrate the framework's effectiveness on the AppWorld benchmark, showing consistent improvements across all difficulty levels, with particularly strong gains on complex tasks where learned experience is most valuable.
\end{itemize}

\section{Problem Statement}

\subsection{The Agent Learning Challenge}

LLM-powered agents execute tasks by iteratively reasoning, selecting actions, and observing outcomes. Each execution trajectory—the complete sequence of thoughts, actions, and results from initial request to final outcome—contains patterns that could inform future executions~\cite{pink2025position}. However, extracting actionable learnings from these trajectories is non-trivial for several reasons.

First, \textbf{valuable patterns exist across diverse outcome categories}. Not all learning opportunities arise from failures. An agent that successfully completes a task may have employed an elegant strategy worth replicating, discovered an efficient API usage pattern, or executed a thorough validation sequence that prevented errors. Conversely, an agent that ultimately succeeds may have done so inefficiently—taking unnecessary steps, making redundant API calls, or using granular operations where bulk operations exist. And agents that encounter failures may successfully recover, with the recovery pattern itself being valuable to capture. A comprehensive learning system must extract insights from clean successes, inefficient successes, failure-then-recovery sequences, and complete failures.

Second, \textbf{causality is often non-obvious from raw logs}. When an agent fails at step 15 of an execution, the problematic decision may have occurred at step 3. When an agent successfully recovers from an error, identifying which specific reasoning led to the recovery requires semantic understanding of the agent's thoughts, not just observation of the final outcome. When an agent completes a task inefficiently, determining which alternative approach would be more efficient requires understanding both what the agent did and what other options were available.

Third, \textbf{learnings must be contextually retrieved}. An optimization tip about using bulk cart operations is relevant when the agent is performing cart management but irrelevant for email composition tasks. A recovery tip about handling authentication failures is critical for tasks involving authenticated APIs but unnecessary for read-only operations. The retrieval system must match learnings to contexts based on multiple dimensions: task type, domain, semantic similarity to current request, and the specific execution patterns involved. The importance of precise retrieval is amplified by empirical evidence that agents closely follow retrieved experiences~\cite{xiong2025memory}, making mismatched retrieval a direct source of degraded performance.

Fourth, \textbf{learnings must be actionable and specific}. Generic advice like "be careful with API calls" provides little value. Effective learnings specify concrete validation checks, particular API usage patterns, specific error recovery sequences, or explicit prerequisite verification steps. They must be formulated in terms the agent can directly apply: "Before initiating checkout, verify payment method is configured by calling \texttt{get\_payment\_methods()} and checking for non-empty results" is actionable; "make sure payment works" is not.

Fifth, \textbf{learnings must be traceable to their source}. Each learning must maintain provenance—a link back to the specific trajectory and outcome from which it was derived~\cite{dechant2025risks}. This enables validation of whether learnings are effective (do similar failures still occur after the learning is deployed?), investigation of why certain guidance was generated, and auditing of the learning system's decisions. Without provenance, it is impossible to debug incorrect guidance, assess learning quality over time, or build trust in the system's recommendations.

\subsection{Learning Requirements}

For agents that reason and act iteratively (e.g., ReAct-style, plan-and-execute), the learning system must satisfy several requirements.

\textbf{Strategy extraction from successful patterns}: When an agent executes a task cleanly—without errors, unnecessary steps, or recovery sequences—its approach often embodies effective strategies. The system must identify these patterns: Did the agent verify prerequisites before attempting operations? Did it systematically explore available APIs before selecting one? Did it validate intermediate results before proceeding to dependent steps? These successful patterns should be encoded as strategy tips that guide future executions toward similarly effective approaches.

\textbf{Recovery extraction from failure handling}: When an agent encounters an error but successfully recovers, the recovery sequence is valuable. The system must identify what went wrong, what the agent recognized about the failure, how it adjusted its approach, and what specific actions led to successful recovery. For example, if an agent attempts checkout without payment configured, receives an error, recognizes the missing payment method, adds payment information, and successfully retries, this entire sequence should be encoded as a recovery tip including the failure pattern, recognition signals, and correction steps.

\textbf{Optimization extraction from inefficient successes}: When an agent successfully completes a task but does so suboptimally, the system must identify the inefficiency and determine the more efficient alternative. This requires understanding not just what the agent did, but what other options were available. For example, if an agent removes items from a cart one-by-one in a loop when a bulk \texttt{empty\_cart()} operation exists, the system must recognize this pattern, identify the more efficient alternative, and encode an optimization tip specifying when and how to use the bulk operation.

\textbf{Step-level decision attribution}: When failures or inefficiencies occur, the system must identify which specific reasoning steps and decisions led to the outcome. This requires semantic analysis of the agent's thoughts, not just observation of actions. If an agent fails because it assumed an API was available without verifying, the attribution must identify the assumption step, explain why it was problematic, and specify what verification should have occurred.

\textbf{Thought pattern recognition}: Agents often exhibit meta-cognitive behaviors that indicate their reasoning quality. An agent that explicitly validates prerequisites is demonstrating a positive pattern. An agent that recognizes its own errors and self-corrects is exhibiting reflection. An agent that makes assumptions without verification is exhibiting a negative pattern. The system must identify these cognitive patterns semantically—recognizing that "I should verify all APIs are available" exhibits a validation pattern even without using the word "validate"—and use them to guide learning extraction.

\textbf{Semantic reasoning analysis}: Beyond recognizing individual thought patterns, the system must move beyond raw execution logs to understand the full structure of agent reasoning. The system must identify and classify distinct reasoning modes—analytical thoughts (examining data or constraints), planning thoughts (formulating action sequences), validation thoughts (checking prerequisites or intermediate results), reflection thoughts (evaluating past actions), and self-correction sequences (recognizing and recovering from errors)—to understand how agents reasoned about tasks and where their reasoning succeeded or failed. This structured understanding of reasoning flows is what enables the extraction of meaningful learnings from trajectories rather than surface-level pattern matching on actions alone.

\subsection{Limitations of Existing Approaches}

Existing approaches to agent improvement fail to address these challenges comprehensively.

\textbf{Rule-based systems} encode decision rules based on anticipated patterns, but they cannot adapt to unforeseen situations and require constant manual maintenance as new patterns emerge. They also cannot automatically extract rules from observed execution trajectories—each rule must be manually crafted by developers who may not have visibility into actual deployment patterns.

\textbf{Prompt engineering} improves agent performance through iteratively refined guidance and examples, but this guidance is generic rather than specific to actual deployment experiences. If an agent repeatedly fails at a particular API authentication flow, prompt engineering might eventually capture this pattern, but only after manual observation and prompt modification. There is no mechanism for automatic improvement based on observed outcomes, and no systematic way to capture the full range of learning opportunities from successes, failures, and recoveries.

\textbf{Generic memory systems} represent a more sophisticated approach but still fall short. Systems like Mem0~\cite{chhikara2025mem0} and Letta~\cite{packer2023memgpt} store facts extracted from conversations in vector databases for later retrieval. However, these systems lack several critical capabilities for agent learning. They have no understanding of agent execution patterns—they treat all memories uniformly rather than distinguishing between strategy patterns, recovery sequences, and optimization opportunities. They cannot perform causal analysis to identify which decisions led to failures or inefficiencies—they store outcomes but not the decision chains that produced them. They lack structured learning extraction with categories, priorities, and actionable steps—memories are typically free-form text without the structure needed for agent guidance. They provide no provenance tracking from learnings back to source trajectories, making it impossible to validate whether learnings are effective or to investigate why certain guidance was generated~\cite{zhang2024survey}.

\textbf{Reinforcement learning approaches} learn from reward signals but have their own limitations for this problem. They require extensive training data to learn effective policies, which may not be available when failures are rare but consequential. They are computationally expensive to train and update, making them impractical for continuously evolving agent systems. They provide limited interpretability regarding why certain decisions improve outcomes—the learned policy is often a black box. For scenarios where understanding the reasoning behind improvements is valuable (such as debugging or auditing agent behavior), RL approaches provide insufficient transparency. Additionally, RL approaches struggle with the multi-category learning problem—they optimize for overall reward but do not naturally distinguish between strategy patterns, recovery sequences, and optimization opportunities.

\section{Approach}

\begin{figure*}
    \includegraphics[width=\textwidth]{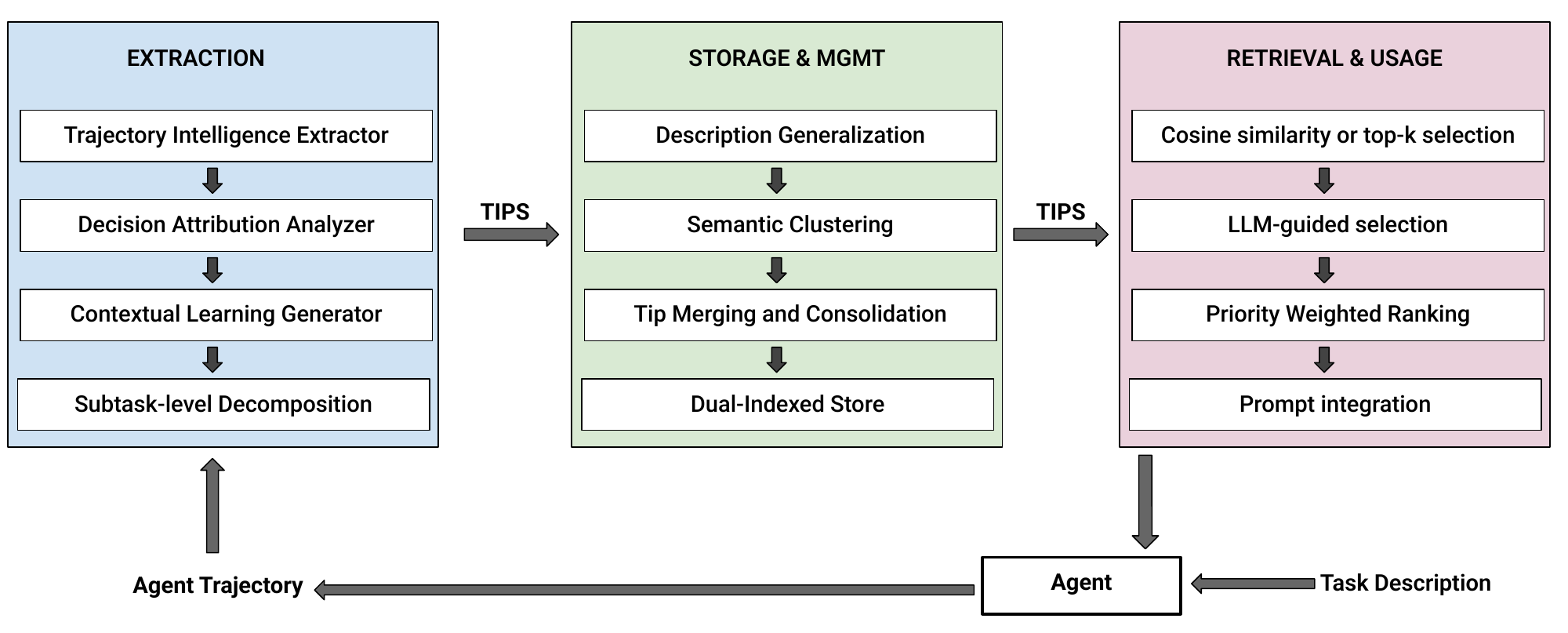}
    \caption{Overview of our approach}~\label{fig:soln-overview}
\end{figure*}

As illustrated in Figure~\ref{fig:soln-overview}, we propose a framework that transforms raw agent execution trajectories into actionable, contextually-retrieved guidance for future invocations. The framework operates as a three-phase pipeline:

\begin{enumerate}
\item \textbf{Phase 1: Trajectory Analysis and Tips Extraction.} Given an agent's execution trajectory for a completed task, the system analyzes the reasoning trace to identify causal decision chains—why outcomes occurred—and extracts structured tips capturing effective strategies, recovery patterns, and optimization opportunities. Tips are extracted at two complementary granularities: \emph{task-level} tips that capture holistic end-to-end patterns, and \emph{subtask-level} tips that decompose trajectories into reusable logical phases (authentication, data retrieval, processing, etc.) for cross-task transfer.

\item \textbf{Phase 2: Tip Storage and Management.} Extracted tips are generalized, clustered, and consolidated before storage. Subtask descriptions are abstracted to remove entity-specific details, enabling semantic clustering of tips from different tasks that share common subtask patterns. An LLM-based merging process consolidates redundant or overlapping tips within each cluster, producing a curated memory of non-redundant, high-quality guidance. Tips are stored with dual representations—vector embeddings for semantic search and structured metadata for filtering.

\item \textbf{Phase 3: Runtime Retrieval.} When an agent is invoked for a new task, the system retrieves relevant tips from memory and injects them into the agent's prompt as guidelines before reasoning begins. Two retrieval strategies are supported: cosine similarity retrieval (fast, no LLM call) and LLM-guided selection (richer reasoning about task context at the cost of an additional LLM invocation).
\end{enumerate}

These phases form a self-reinforcing cycle: as more trajectories are processed, the memory system accumulates increasingly comprehensive and refined guidance. Agents that receive this guidance produce higher-quality trajectories that may reveal subtler patterns for further learning. The following subsections detail each phase.

\subsection{Phase 1: Trajectory Analysis and Tips Extraction}

This phase analyzes completed agent trajectories to extract structured, actionable tips. It comprises three stages: trajectory intelligence extraction, decision attribution analysis, and tip generation. A key design dimension of the tip generation stage is the \emph{granularity} at which tips are extracted—either at the level of entire task trajectories (\emph{task-level}) or at the level of individual logical subtasks within a trajectory (\emph{subtask-level}). We explore both granularities and compare their effectiveness in Section~\ref{sec:evaluation}.

\subsubsection{Trajectory Intelligence Extractor}

The Trajectory Intelligence Extractor transforms raw agent execution data into a structured intermediate representation that captures semantic meaning beyond what traditional logging provides: why agents made particular decisions, how they validated their reasoning, where they exhibited self-corrective behavior, and what patterns characterized successful versus unsuccessful executions.

The component receives raw agent trajectories containing sequential steps with agent invocations, prompts or contexts, agent responses including thoughts and reflections, actions taken and their results, and optionally, evaluation reports or ground-truth outcome assessments. Each trajectory represents a complete task execution from initial user request through final outcome. Crucially, ground-truth outcome labels (success or failure) are not required: when they are available—for instance, from a benchmark evaluation harness—the system uses them directly to classify the trajectory; when they are absent, the system infers outcome from the agent's own self-reflective signals identified in subsequent stages.

The first processing stage parses agent responses to identify and categorize reasoning into four types based on cognitive function:
\textbf{Analytical thoughts} where the agent analyzes the situation and assesses constraints;
\textbf{Planning thoughts} where the agent decides what actions to take and in what sequence;
\textbf{Validation thoughts} where the agent checks assumptions or verifies preconditions; and
\textbf{Reflection thoughts} where the agent reconsiders its approach, often triggered by unexpected results.
Beyond categorization, the extractor identifies status indicators, execution summaries, and error recognition statements, enabling downstream components to understand the reasoning process that led to actions.

The second stage uses an LLM to identify cognitive patterns within extracted thoughts through semantic understanding rather than keyword matching. The system recognizes:
\textbf{Validation patterns}—any expression of checking or verifying assumptions, even without validation-related keywords (e.g., ``I need to ensure all required APIs are included'' exhibits validation behavior);
\textbf{Reflection patterns}—reconsideration of previous decisions, often after errors;
\textbf{Self-correction patterns}—proactively identifying and fixing errors before external signals;
\textbf{Error recognition patterns}—noticing problems that may affect task completion;
\textbf{API discovery patterns}—systematic exploration of available APIs; and
\textbf{Efficiency awareness patterns}—considering whether more efficient alternatives exist.
This semantic approach generalizes across linguistic variations, unlike rule-based keyword matching.

The third stage determines the trajectory outcome. When ground-truth evaluation reports are present, the stage interprets them with semantic understanding: a report stating ``API response returned 400 Bad Request'' is converted into ``Checkout API failed because required payment method parameter was not provided,'' and for each outcome indicator, the module determines what the test validates, why it failed (if applicable), the impact on task completion, and overall quality assessment. When ground-truth labels are absent, the stage instead synthesizes outcome from the self-reflective signals extracted in stages 1 and 2—reflection thoughts, self-correction patterns, and error recognition patterns—to infer whether the agent succeeded, failed, or recovered. In both cases, the result is an outcome classification used by downstream components.

A fourth stage specifically analyzes successful executions, distinguishing:
\textbf{Clean successes}—task completed without errors or unnecessary steps, with patterns that are candidates for strategy tips;
\textbf{Inefficient successes}—task completed but suboptimally (e.g., repeated operations that could be batched), yielding candidates for optimization tips; and
\textbf{Recovery sequences}—successful error handling within otherwise successful executions, yielding candidates for recovery tips.

The output is a structured intermediate representation enriched with extracted thoughts, identified patterns with confidence scores, evaluation intelligence, success patterns, and metadata including trajectory identifier, task intent, step count, and overall outcome classification.

\subsubsection{Decision Attribution Analyzer}

The Decision Attribution Analyzer performs automated causal analysis to identify which decisions and reasoning steps led to observed outcomes. It analyzes all outcome types—not just failures.

The first stage scans the intermediate representation for outcome indicators across four categories:
\textbf{Failure indicators}—failed evaluations, error messages, task incompletion signals;
\textbf{Recovery indicators}—failure followed by successful completion, error recognition followed by corrective actions;
\textbf{Inefficiency indicators}—repeated operations that could be batched, unnecessary intermediate steps, granular operations where bulk alternatives exist; and
\textbf{Success patterns}—clean completion, systematic prerequisite verification, efficient API usage.
For each detected outcome, contextual information is extracted as the starting point for causal analysis. Importantly, the outcome location is typically not the cause location.

The causal analysis module uses an LLM to trace backwards through the agent's reasoning steps to identify which decisions led to the observed outcome.
For \textbf{failures}, the analysis distinguishes: the \emph{immediate cause} (what directly triggered the failure), the \emph{proximate cause} (recent decisions that enabled it), the \emph{root cause} (the underlying issue that originated the chain), and \emph{contributing factors}.
For \textbf{recoveries}, it identifies what enabled the failure, how the agent recognized the problem, what corrective action was taken, and why the correction succeeded.
For \textbf{inefficiencies}, it identifies what made execution suboptimal, what more efficient alternative exists, why the alternative is better, and whether the agent was aware of the inefficiency.
For \textbf{success patterns}, it identifies what strategies contributed to clean success, why they were effective, and what made the approach particularly good.

The final stage generates specific prevention or improvement steps for each attributed decision point. These steps must be
\textbf{actionable}—the agent can actually perform them;
\textbf{specific}—concrete actions rather than vague advice;
\textbf{causal}—directly addressing the root cause; and
\textbf{preventive or improving}—stopping similar failures from occurring or specifying more efficient approaches.

\subsubsection{Contextual Learning Generator}
\label{sec:learning-generator}

The Contextual Learning Generator converts decision analyses into reusable memory entries that are actionable, contextually rich, and properly categorized. The key innovation is generating three distinct tip types based on trajectory outcomes.

\textbf{Strategy tips} encode effective patterns from clean successful executions—what worked well and should be replicated. Example:
\begin{lstlisting}
Content: "When performing checkout operations, systematically verify all prerequisites (cart contents, shipping address, payment method) before initiating the checkout sequence."
Category: strategy
Steps:
1. Call get_cart_items() to verify cart is not empty
2. Call get_shipping_address() to verify address is configured
3. Call get_payment_methods() to verify payment method exists
4. Only proceed with checkout if all prerequisites are satisfied
Trigger: "When task involves checkout, purchase, or payment operations"
\end{lstlisting}

\textbf{Recovery tips} encode both the failure pattern and the recovery pattern from failure-then-recovery sequences. Example:
\begin{lstlisting}
Content: "When checkout fails with 'payment method required' error, verify payment configuration and add payment method if missing before retrying."
Category: recovery
Steps:
1. Recognize error message indicating missing payment method
2. Call get_payment_methods() to check current configuration
3. If empty, call add_payment_method() with appropriate details
4. Retry the checkout operation
Trigger: "When checkout or payment operations fail"
Negative Example: "Do not simply retry without addressing the missing payment method."
\end{lstlisting}

\textbf{Optimization tips} identify efficiency improvements from successful but suboptimal executions. Example:
\begin{lstlisting}
Content: "When emptying a shopping cart with multiple items, use empty_cart() instead of iterating remove_from_cart(item_id) for each item."
Category: optimization
Steps:
1. Check if cart has multiple items to remove
2. Instead of looping remove_from_cart(), call empty_cart() once
3. Verify cart is empty with get_cart_items()
Trigger: "When task requires removing all items from cart"
Negative Example: "Do not use for i in items: remove_from_cart(i) when emptying the entire cart."
\end{lstlisting}

The system analyzes trajectories to determine contextual dimensions for both generation and retrieval: the \emph{application context} (which domain the task involves), the \emph{task category} (type of operation within the domain), and the \emph{complexity level}. Tip content is generated using specialized prompts for each category, incorporating the relevant execution patterns, and each prompt includes guidelines for generating actionable, specific, generalizable tips.

Each generated memory entry contains: a unique identifier, tip category (strategy, recovery, optimization), actionable content, explanatory purpose, concrete implementation steps, trigger condition, optional negative example, application context (or null for generic tips), task category (or null for generic tips), priority level (critical/high/medium/low based on outcome severity), source trajectory ID, and source outcome description.

The system also generates both domain-specific and generic tips from the same trajectory, maximizing precision and coverage. From a failure involving missing payment APIs in e-commerce checkout, the system generates a domain-specific tip (``For e-commerce tasks involving checkout, verify payment method is configured before initiating checkout'') and a generic tip (``When initiating operations that have prerequisites, systematically verify all prerequisites before beginning''). This dual-level generalization ensures high precision when context matches domain-specific tips and broad coverage through generic tips that apply even in novel domains.

\subsubsection{Task-Level vs.\ Subtask-Level Extraction}
\label{sec:subtask-extraction}

The tip generation stage can operate at two granularities. \emph{Task-level} extraction treats an entire trajectory as a unit, producing holistic tips that capture end-to-end execution patterns. \emph{Subtask-level} extraction first decomposes the trajectory into logical subtasks and then extracts focused tips for each subtask independently.

The two approaches offer different tradeoffs. Task-level tips are straightforward to extract and capture overarching strategies spanning the full task. However, their reusability is limited by task specificity: a tip extracted from ``Name the artist most recommended to me on Spotify'' may not transfer to ``Move my go-to-sleep phone alarm to 20 minutes later,'' even though both share common subtasks such as authentication and paginated data retrieval. Task-level tips also bundle concerns from distinct execution phases, reducing retrieval precision.

Subtask-level extraction addresses these limitations by scoping each tip to a single logical phase. Many tasks share common subtasks that generalize across contexts:

\begin{itemize}
\item \textbf{Authentication subtasks} follow a common pattern across apps (Spotify, Phone, Venmo): retrieve credentials from a supervisor, login, and store the access token.
\item \textbf{Data retrieval subtasks} share pagination patterns: issue paginated API calls, aggregate results, and store them for downstream processing.
\item \textbf{Data processing subtasks} involve domain-independent operations: counting, filtering, aggregation, and transformation of retrieved data.
\item \textbf{Task completion subtasks} are near-universal: reporting results and marking tasks complete.
\end{itemize}

By extracting tips at this granularity, we enable cross-task transfer (authentication tips from Spotify tasks help with Phone app tasks), better matching (a task about updating alarms retrieves tips from a ``retrieve all alarms'' subtask even if the original task was about deleting alarms), and compositional learning (new complex tasks leverage tips from multiple simpler subtasks).

\paragraph{Two-Phase Extraction Pipeline.}
The subtask-level extraction operates as a two-phase LLM-based pipeline.

\textbf{Phase A: Trajectory Segmentation.} An LLM analyzes the full agent trajectory and segments it into logical subtasks. For each subtask, the model produces a generalized description (deliberately generic, e.g., ``Authenticate with Spotify'' rather than ``Login as user@gmail.com''), the set of applications involved, the step range in the original trajectory (maintaining traceability), and the subtask's purpose. The segmentation prompt instructs the model to identify natural boundaries between distinct logical phases—transitions from authentication to data retrieval, from data retrieval to processing, and so on.

For example, a trajectory for ``Name the artist most recommended to me on Spotify'' might be segmented into: (1) discover relevant APIs and their specifications, (2) authenticate with Spotify, (3) retrieve recommended songs via paginated requests, and (4) analyze recommendations to determine the most recommended artist.

\textbf{Phase B: Per-Subtask Tips Extraction.} An LLM independently extracts 2--4 actionable tips for each subtask. By scoping each extraction call to a single subtask, the prompts remain focused and the tips avoid conflating concerns from different execution phases. The tips are constrained to be concrete (specific API patterns rather than vague advice), generalizable (avoiding task-specific details such as particular email addresses, song names, or payment amounts), and actionable (directly applicable by an agent encountering a similar subtask). Optionally, different models can be used for Phase A and Phase B—a more capable model for segmentation and a lighter model for per-subtask extraction—to balance cost and quality.

Example output for the ``Authenticate with Spotify'' subtask:
\begin{lstlisting}
Tips:
1. "Always retrieve account credentials from supervisor.show_account_passwords() before attempting authentication"
2. "Immediately store and validate access tokens after login to ensure successful subsequent API calls"
3. "Filter credentials by app name to select the correct password for the target service"
\end{lstlisting}

Subtask-level and task-level tips are complementary rather than competing. Task-level tips capture holistic patterns about end-to-end execution strategy (e.g., ``verify all prerequisites before checkout''), while subtask-level tips capture focused patterns about specific execution phases (e.g., ``use paginated retrieval when fetching large result sets''). Both levels are stored in the same memory system and can be retrieved together during Phase~3.

\subsection{Phase 2: Tip Storage and Management}

As tips accumulate from many trajectories across diverse tasks, the memory system must address redundancy, inconsistency, and scalability. Two trajectories involving e-commerce checkout may independently produce tips about verifying payment methods; dozens of trajectories across different apps will produce authentication-related tips with overlapping guidance. Without consolidation, the memory grows linearly with the number of processed trajectories, retrieval quality degrades as near-duplicate tips compete for limited prompt space, and contradictory tips from different trajectories may confuse the agent.

Phase~2 addresses these challenges through a pipeline of subtask description generalization, semantic clustering, and LLM-based tip consolidation.

\subsubsection{Subtask Description Generalization}

Subtask descriptions produced by Phase~1 contain varying levels of specificity that hinder clustering. ``Retrieve Spotify password for john.doe@email.com using supervisor API,'' ``Get Venmo login credentials for user alice\_smith,'' and ``Fetch Phone app password from supervisor'' all describe the same abstract operation: retrieving service credentials. To enable meaningful clustering, the system generalizes subtask descriptions through three transformations:

\begin{itemize}
\item \textbf{Entity abstraction}: Replaces specific user names, email addresses, app names, item IDs, and other entity references with generic placeholders. ``Retrieve Spotify password for john.doe@email.com'' becomes ``Retrieve service account credentials.''
\item \textbf{Action normalization}: Maps semantically equivalent verbs and phrases to canonical forms. ``Get,'' ``fetch,'' ``retrieve,'' and ``obtain'' are normalized to a single canonical verb. ``Log in,'' ``sign in,'' and ``authenticate'' are similarly unified.
\item \textbf{Context removal}: Strips task-specific contextual qualifiers that do not affect the subtask's core operation. ``Retrieve credentials in order to check subscription status'' is reduced to ``Retrieve service account credentials,'' since the downstream purpose does not change how credential retrieval should be performed.
\end{itemize}

These transformations are applied using an LLM with a prompt that instructs it to produce maximally abstract descriptions while preserving the core operation. The generalized descriptions serve as the basis for clustering: tips whose generalized subtask descriptions are semantically similar are likely to contain overlapping or complementary guidance.

\subsubsection{Semantic Clustering}

The system clusters tips by computing cosine similarity between the vector embeddings of their generalized subtask descriptions, then applying hierarchical agglomerative clustering with a similarity threshold. Two generalized descriptions such as ``Retrieve service account credentials'' and ``Authenticate with external service'' may describe distinct subtasks despite surface-level relatedness, while ``Retrieve service account credentials'' and ``Obtain application login credentials'' describe the same operation. Hierarchical clustering with an appropriate threshold (empirically, $\sim$0.85 on generalized descriptions) groups truly equivalent subtask descriptions while keeping distinct operations separate.

Within each cluster, all associated tips are collected regardless of their source trajectory, task context, or extraction granularity. A cluster for ``Retrieve service account credentials'' might contain tips from Spotify authentication trajectories, Venmo login trajectories, and Phone app credential retrieval—all reflecting the same underlying subtask pattern observed across different tasks.

\subsubsection{Tip Consolidation and Merging}

Within each cluster, an LLM-based consolidation process merges redundant tips, resolves conflicts, and produces a curated set of non-overlapping guidance. The consolidation operates in three steps:

\textbf{Deduplication.} Tips with near-identical content are identified and merged. ``Always call \texttt{show\_account\_passwords()} before login'' and ``Retrieve credentials using the supervisor password API before authentication'' convey the same guidance; the consolidation produces a single canonical tip that captures the shared insight.

\textbf{Conflict resolution.} When tips from different trajectories offer contradictory guidance (e.g., one tip recommends retrying failed authentication immediately while another recommends re-retrieving credentials first), the system uses outcome metadata—tip category, priority level, and source trajectory success/failure status—to determine which guidance is more reliable. Tips derived from successful trajectories take precedence over those from failed ones, and recovery tips that encode proven correction patterns take precedence over speculative prevention strategies.

\textbf{Synthesis.} Complementary tips that address different aspects of the same subtask are synthesized into coherent, comprehensive guidance. If one tip covers credential retrieval and another covers token validation after login, the consolidated output combines both into a single tip with ordered steps covering the full authentication workflow.

The consolidation also produces a \emph{canonical cluster description}—a single generalized subtask description that represents the cluster for retrieval purposes. This description is re-embedded and stored alongside the consolidated tips, replacing the individual per-trajectory descriptions.

\subsubsection{Storage Representation}

Each consolidated memory entry is stored with two complementary representations. The \textbf{vector embedding} is a dense vector computed from the tip content and purpose using a text embedding model. This captures semantic meaning, enabling similarity search across different terminology—for instance, a tip about ``renewing a subscription'' can match a task description mentioning ``extending my membership,'' and a tip about ``scheduling a recurring event'' can match ``set up a weekly meeting.'' The \textbf{structured metadata} consists of filterable attributes: tip category (strategy, recovery, optimization), priority level, application context, task category, source trajectory IDs (plural, since consolidated tips may derive from multiple trajectories), and creation timestamp.

Tips are indexed by their canonical cluster description for subtask-level tips, and by the original task description for task-level tips, creating natural groupings that enable retrieval at both granularities.

\subsection{Phase 3: Runtime Retrieval}

When an agent is invoked to execute a new task with description $d$, the system retrieves relevant tips from memory and injects them into the agent's prompt as guidelines before reasoning begins. The retrieval strategy directly affects whether the agent receives relevant, actionable guidance or is distracted by irrelevant tips. We consider two strategies with different cost-accuracy tradeoffs.

\subsubsection{Cosine Similarity Retrieval}

The most straightforward approach embeds the incoming task description $d$ and computes cosine similarity against the embeddings of stored task (and subtask) descriptions. Tips associated with the most similar stored descriptions are retrieved and injected into the prompt. This strategy requires no LLM calls at retrieval time and is fast and inexpensive—a pure vector database lookup.

Two complementary mechanisms control which tips are selected:

\begin{itemize}
\item \textbf{Similarity threshold $\tau$}: Only tips whose source description has cosine similarity $\geq \tau$ with $d$ are eligible. A \emph{high threshold} (e.g., $\tau \geq 0.85$) ensures retrieved tips are closely related to the current task, but risks excluding tips from tasks that are semantically equivalent yet phrased differently. For example, ``I want an Amazon Prime membership'' and ``Sign me up for Amazon Prime'' describe the same task but may have cosine similarity below 0.85 due to lexical differences. A \emph{low threshold} (e.g., $\tau \leq 0.6$) casts a wider net, but risks pulling in tips from unrelated tasks—tips from ``Book a flight to New York'' are unlikely to help an agent executing ``Update my calendar for next week,'' yet both involve scheduling-adjacent language that could produce moderate similarity scores.

\item \textbf{Top-$k$ selection}: After filtering by threshold, the system selects the $k$ highest-scoring tips. This bounds the number of tips injected into the prompt regardless of how many pass the threshold, preventing prompt bloat when many stored tasks are moderately similar.
\end{itemize}

In practice, these two mechanisms are combined: the system retrieves all tips with similarity $\geq \tau$, then selects the top $k$ by similarity score. Typical values are $\tau \in [0.5, 0.7]$ and $k \in [5, 10]$.

\subsubsection{LLM-Guided Selection}

A more expressive approach uses an LLM at retrieval time to analyze the task description $d$, detect the application context and task category, and reason about which types of guidance are most relevant. The LLM constructs a structured retrieval query that combines:

\begin{itemize}
\item \textbf{Metadata filters}: The LLM identifies that a task about ``Complete my pending Venmo payment requests'' involves the Venmo application and payment operations, and constrains retrieval to tips from the payment domain (or generic tips with null application context).
\item \textbf{Category awareness}: Based on the task description, the LLM may determine that recovery tips are particularly relevant (e.g., the task mentions retrying a failed payment) or that strategy tips should be prioritized (e.g., the task involves a multi-step workflow).

\end{itemize}

LLM-guided selection is more expensive (requiring an additional LLM call per task) but can reason about nuances that pure embedding similarity misses. For instance, an LLM can recognize that ``Delete all my read emails older than 30 days'' and ``Clean up my inbox by removing old messages'' are the same task even when their embeddings diverge, and it can infer that a task involving ``checkout'' implies payment-related tips are relevant even if ``payment'' is never mentioned in the task description.

\textbf{Comparison.} Cosine similarity retrieval is simple, fast, and requires no LLM calls at runtime—making it suitable for latency-sensitive or cost-constrained deployments. LLM-guided selection provides richer reasoning about task context at the cost of an additional LLM invocation. We evaluate both strategies empirically in Section~\ref{sec:evaluation}.

\subsubsection{Prompt Integration}

Regardless of retrieval strategy, the selected tips are injected into the agent's prompt as a ``guidelines'' section positioned after the task context but before the standard agent instructions. Each tip is formatted to be quickly scannable and actionable, highlighting priority level, category, actionable content, purpose, implementation steps, and trigger condition. For example:

\begin{lstlisting}
[PRIORITY: HIGH] Recovery Tip:
When a login attempt fails with "invalid credentials," verify you are using the correct app-specific password by re-calling supervisor.show_account_passwords() and filtering by the target app name.

Apply when: Authentication fails on any app after an initial login attempt.
Steps:
1. Re-retrieve credentials from supervisor
2. Filter for the specific app name (exact match)
3. Retry login with the correct credentials
\end{lstlisting}

This formatting enables agents to quickly identify the type of guidance, prioritize critical tips, and understand both what to do and why. The prompt integration creates a feedback loop: agents receiving relevant tips avoid failure patterns, execute more efficiently, and apply successful strategies, producing higher-quality trajectories that reinforce the memory system's value.

\section{Evaluation}
\label{sec:evaluation}

We evaluate our trajectory-informed memory generation framework on the AppWorld benchmark, a comprehensive evaluation suite for LLM agents that perform complex tasks across multiple applications. Our evaluation examines two dimensions: (1) the effect of tip extraction granularity (task-level vs.\ subtask-level tips), and (2) the effect of retrieval strategy (cosine similarity vs.\ LLM-guided selection). The evaluation demonstrates that agents equipped with learned memory from past executions substantially outperform agents without memory, with particularly strong improvements on challenging tasks.

\subsection{Experimental Setup}

\subsubsection{Benchmark Description}

AppWorld is a benchmark designed to evaluate LLM agents on realistic task completion across diverse application domains. The benchmark contains tasks spanning e-commerce, email, calendar, file management, and other common application scenarios. Each task consists of a natural language instruction that the agent must execute by interacting with APIs provided for various applications.

The benchmark includes two key evaluation metrics:

\textbf{Task Goal Completion (TGC)} measures the percentage of individual tasks where the agent passes all programmatic unit tests, which verify correct API usage, database state changes, and expected end states. Each task is a complex, multi-step, app-based challenge that typically requires multiple API calls across an average of 1.8 apps and 9.5 APIs. A task is successful only if all unit tests pass.

\textbf{Scenario Goal Completion (SGC)} measures the percentage of scenarios where the agent correctly completes \textit{all} task variants (typically three) associated with a given scenario, testing for consistency across related tasks. A scenario is only counted as successful if every variant passes, making this a stricter metric than TGC.

Tasks in AppWorld are categorized by difficulty level:
\begin{itemize}
\item \textbf{Difficulty 1 (Easy)}: Simple tasks requiring basic API interactions, typically single-domain with straightforward execution sequences
\item \textbf{Difficulty 2 (Medium)}: Moderate complexity tasks that may span multiple domains or require conditional logic and error handling
\item \textbf{Difficulty 3 (Hard)}: Complex multi-step tasks requiring careful planning, prerequisite management, cross-domain coordination, and robust error recovery, often involving 50+ lines of equivalent code and up to 26 APIs
\end{itemize}

\subsubsection{Agent Configuration}

We evaluate using a single-agent configuration implementing a simplified ReAct-style reasoning and action loop. The agent iteratively reasons about the current task state, selects actions to take, executes those actions via API calls, and observes the results. The agent continues this loop until it determines the task is complete or encounters an unrecoverable failure. Both the agent and the tip extraction pipeline use GPT-4.1.

The base agent (without memory) receives only the task instruction and standard prompting that includes its role description, available APIs, and general guidelines for task execution. The memory-enhanced agent additionally receives retrieved tips from the memory system, injected into the prompt before the agent begins reasoning.

\subsubsection{Tip Extraction Configurations}

We evaluate two tip extraction granularities:

\textbf{Task-level tips} are extracted from entire trajectories as described in Section~\ref{sec:learning-generator}. Each trajectory produces a holistic set of strategy, recovery, and optimization tips that capture end-to-end execution patterns. Task-level tips are well-suited for capturing overarching strategies (e.g., ``verify all prerequisites before checkout'') but may bundle unrelated concerns from different execution phases.

\textbf{Subtask-level tips} are extracted using the two-phase pipeline described in Section~\ref{sec:subtask-extraction}. Trajectories are first segmented into logical subtasks (authentication, data retrieval, data processing, etc.), and tips are then extracted independently for each subtask. Subtask-level tips are more focused and reusable across tasks that share common subtasks.

Both tip types were generated from agent executions on the AppWorld training and development partitions, processed through our pipeline.

\subsubsection{Retrieval Strategy Configurations}

We evaluate two retrieval strategies for selecting which tips to inject into the agent's prompt at runtime:

\textbf{Cosine similarity retrieval} embeds the task instruction using a text embedding model and retrieves the top-$k$ tips whose vector embeddings have the highest cosine similarity to the query embedding. This is a standard retrieval approach that requires no LLM calls at retrieval time and is fast and inexpensive.

\textbf{LLM-guided selection} uses an LLM to analyze the task instruction, detect the application context and task category, and construct a retrieval query that combines semantic similarity with metadata filtering and priority-weighted ranking. This approach is more expressive—it can reason about which tip categories are most relevant and ensure critical tips surface—but requires an additional LLM call at retrieval time.

For both strategies, the top 5 tips are retrieved and injected into the agent's prompt before reasoning begins.

\subsubsection{Evaluation Protocol}

We evaluated configurations on three partitions of AppWorld: (1) the \textbf{test-normal} partition, which contains held-out tasks not seen during memory generation, measuring the agent's ability to generalize learned patterns to novel tasks; (2) the \textbf{train} partition, from which tips were generated, measuring how effectively tips improve performance when the same task is encountered again; and (3) the \textbf{dev} partition, also used during tip generation, providing a complementary view.

Each task was executed independently with a maximum of 30 reasoning-action steps. Task and scenario goal completion were assessed using AppWorld's automated evaluation framework, which verifies that all explicit requirements (task goals) and implicit requirements (scenario goals) are satisfied by examining the final state of all involved applications.

\subsection{Held-Out Results (Test-Normal)}

The test-normal partition contains tasks not seen during memory generation, providing the most rigorous evaluation of the memory system's ability to generalize learned patterns to novel tasks. We present results for multiple configurations.

\subsubsection{Subtask-Level Tips with LLM-Guided Selection}

Tables~\ref{tab:test-mem} and~\ref{tab:test-base} present results for subtask-level tips with LLM-guided selection---the best-performing configuration for scenario goal completion.

\begin{table}[h]
\centering
\caption{Subtask Tips + LLM Selection: Test-Normal}
\label{tab:test-mem}
\begin{tabular}{l|c|c}
\hline
\textbf{Type} & \textbf{Task Goal} & \textbf{Scenario Goal} \\
\hline
Aggregate & 73.2 & 64.3 \\
Difficulty 1 & 91.2 & 89.5 \\
Difficulty 2 & 70.8 & 56.2 \\
Difficulty 3 & 58.7 & 47.6 \\
\hline
\end{tabular}
\end{table}

\begin{table}[h]
\centering
\caption{Baseline Agent (No Memory): Test-Normal}
\label{tab:test-base}
\begin{tabular}{l|c|c}
\hline
\textbf{Type} & \textbf{Task Goal} & \textbf{Scenario Goal} \\
\hline
Aggregate & 69.6 & 50.0 \\
Difficulty 1 & 89.5 & 79.0 \\
Difficulty 2 & 66.7 & 56.2 \\
Difficulty 3 & 54.0 & 19.1 \\
\hline
\end{tabular}
\end{table}

The memory-enhanced agent achieves 73.2\% TGC compared to 69.6\% for the baseline (+3.6~pp) and 64.3\% SGC compared to 50.0\% (+14.3~pp). The larger SGC improvement suggests that the memory system not only helps agents complete individual tasks correctly but substantially improves consistency across task variants within scenarios. Since SGC requires all variants to pass, it is sensitive to sporadic failures---exactly the brittleness that learned tips help mitigate.

The benefits scale with task complexity. Difficulty~1 tasks show improvements of +1.7~pp TGC and +10.5~pp SGC, with the baseline already achieving high TGC. Difficulty~2 tasks show +4.1~pp TGC with no SGC change, benefiting from learned patterns around error handling and prerequisite verification. Difficulty~3 tasks show the most dramatic improvements: +4.7~pp on TGC and a remarkable +28.5~pp on SGC (19.1\% $\rightarrow$ 47.6\%), a 149\% relative increase. These complex tasks require sophisticated planning and robust error recovery---areas where the memory system provides the most guidance.

\subsubsection{Task-Level Tips with Cosine Similarity Retrieval}

We next evaluate task-level tips with cosine similarity retrieval. Task-level tips extract holistic insights from entire trajectories rather than decomposing them into subtasks. At retrieval time, the incoming task description is embedded and compared against stored task description embeddings; tips from descriptions exceeding a similarity threshold $\tau$ are retrieved. We evaluate three retrieval parameter configurations to examine the effect of threshold and top-$k$ selection.

\begin{table}[h]
\centering
\caption{Task-Level Tips + Cosine ($\tau \geq 0.5$, top-3): Test-Normal}
\label{tab:task-cos-05-top3}
\begin{tabular}{l|c|c}
\hline
\textbf{Type} & \textbf{Task Goal} & \textbf{Scenario Goal} \\
\hline
Aggregate & 66.7 & 48.2 \\
Difficulty 1 & 86.0 & 68.4 \\
Difficulty 2 & 70.8 & 56.2 \\
Difficulty 3 & 46.0 & 23.8 \\
\hline
\end{tabular}
\end{table}

\begin{table}[h]
\centering
\caption{Task-Level Tips + Cosine ($\tau \geq 0.6$): Test-Normal}
\label{tab:task-cos-06}
\begin{tabular}{l|c|c}
\hline
\textbf{Type} & \textbf{Task Goal} & \textbf{Scenario Goal} \\
\hline
Aggregate & 72.0 & 62.5 \\
Difficulty 1 & 91.2 & 84.2 \\
Difficulty 2 & 72.9 & 68.8 \\
Difficulty 3 & 54.0 & 38.1 \\
\hline
\end{tabular}
\end{table}

\begin{table}[h]
\centering
\caption{Task-Level Tips + Cosine ($\tau \geq 0.5$): Test-Normal}
\label{tab:task-cos-05}
\begin{tabular}{l|c|c}
\hline
\textbf{Type} & \textbf{Task Goal} & \textbf{Scenario Goal} \\
\hline
Aggregate & 70.2 & 57.1 \\
Difficulty 1 & 91.2 & 84.2 \\
Difficulty 2 & 64.6 & 43.8 \\
Difficulty 3 & 55.6 & 42.9 \\
\hline
\end{tabular}
\end{table}

The three cosine similarity configurations reveal important interactions between threshold, top-$k$ selection, and task complexity.

\textbf{Top-$k$ restriction hurts performance.} The most restrictive configuration ($\tau \geq 0.5$, top-3) performs \emph{below} the baseline at the aggregate level (66.7\% TGC, 48.2\% SGC), a drop of $-2.9$~pp and $-1.8$~pp respectively. The top-3 restriction limits the agent to tips from only three matched task descriptions, which may exclude relevant guidance. This is especially damaging for complex tasks: Difficulty~3 drops to 46.0\% TGC ($-8.0$~pp from baseline).

\textbf{Threshold $\tau = 0.6$ is the sweet spot.} The configuration with $\tau \geq 0.6$ (no top-$k$ restriction) achieves the strongest overall results among cosine similarity configurations: 72.0\% TGC (+2.4~pp) and 62.5\% SGC (+12.5~pp). This threshold strikes an effective balance: tight enough to exclude tips from unrelated tasks, yet loose enough to capture semantically equivalent task descriptions that differ lexically (e.g., ``I want an Amazon Prime membership'' and ``Sign me up for Amazon Prime''). The Difficulty~3 SGC improvement is striking: 19.1\% $\rightarrow$ 38.1\% (+19.0~pp), a 99\% relative increase.

\textbf{Lower threshold includes noise.} Dropping to $\tau \geq 0.5$ (no top-$k$) yields 70.2\% TGC (+0.6~pp) and 57.1\% SGC (+7.1~pp)---better than the baseline but weaker than $\tau \geq 0.6$ on both metrics. The lower threshold admits tips from marginally related tasks, diluting the signal. Interestingly, Difficulty~3 TGC is slightly higher with $\tau \geq 0.5$ (55.6\%) than with $\tau \geq 0.6$ (54.0\%), suggesting that for the most complex tasks, casting a wider net occasionally surfaces useful tips from loosely related tasks. However, the reverse pattern holds for Difficulty~2 (64.6\% vs.\ 72.9\%), where the noise from irrelevant tips is more damaging.

\subsubsection{Subtask-Level Tips with Cosine Similarity Retrieval}

To isolate the effect of the retrieval strategy from the effect of tip granularity, we also evaluate subtask-level tips with cosine similarity retrieval ($\tau \geq 0.6$, no top-$k$)---the same retrieval parameters as the best task-level cosine configuration, but with subtask-level tips instead.

\begin{table}[h]
\centering
\caption{Subtask Tips + Cosine ($\tau \geq 0.6$): Test-Normal}
\label{tab:subtask-cos-06}
\begin{tabular}{l|c|c}
\hline
\textbf{Type} & \textbf{Task Goal} & \textbf{Scenario Goal} \\
\hline
Aggregate & 73.8 & 57.1 \\
Difficulty 1 & 91.2 & 73.7 \\
Difficulty 2 & 72.9 & 56.2 \\
Difficulty 3 & 58.7 & 42.9 \\
\hline
\end{tabular}
\end{table}

This configuration achieves 73.8\% TGC (+4.2~pp over baseline)---the highest TGC of any configuration---and 57.1\% SGC (+7.1~pp). Comparing with subtask-level tips with LLM-guided selection (Table~\ref{tab:test-mem}) isolates the effect of the retrieval strategy while holding tip granularity constant: TGC is slightly higher with cosine retrieval (73.8\% vs.\ 73.2\%), but SGC drops substantially (57.1\% vs.\ 64.3\%, a 7.2~pp gap). This divergence is most pronounced on Difficulty~3, where SGC drops from 47.6\% to 42.9\%. The LLM-guided selection's ability to reason about task context and prioritize tip categories appears critical for cross-variant consistency, even though simple cosine retrieval suffices (and marginally excels) for individual task completion.

\subsubsection{Configuration Comparison}

Table~\ref{tab:retrieval-strategies} compares all configurations on the held-out test-normal partition, using $\tau \geq 0.6$ (no top-$k$) for both cosine similarity configurations.

\begin{table*}[h]
\centering
\caption{Configuration Comparison on Test-Normal (Aggregate)}
\label{tab:retrieval-strategies}
\begin{tabular}{l|l|c|c|c|c}
\hline
\textbf{Tip Granularity} & \textbf{Retrieval Strategy} & \textbf{TGC} & \textbf{$\Delta$ TGC} & \textbf{SGC} & \textbf{$\Delta$ SGC} \\
\hline
\multicolumn{2}{l|}{\textit{Baseline (no memory)}} & 69.6 & --- & 50.0 & --- \\
\hline
Subtask-level & LLM-guided selection & 73.2 & +3.6 & \textbf{64.3} & \textbf{+14.3} \\
Subtask-level & Cosine sim.\ ($\tau \geq 0.6$) & \textbf{73.8} & \textbf{+4.2} & 57.1 & +7.1 \\
Task-level & Cosine sim.\ ($\tau \geq 0.6$) & 72.0 & +2.4 & 62.5 & +12.5 \\
\hline
\end{tabular}
\end{table*}

The three configurations reveal a clear separation between what drives task goal completion versus scenario goal completion.

\textbf{Tip granularity drives TGC.} Subtask-level tips outperform task-level tips on TGC regardless of retrieval strategy: 73.8\% (cosine) and 73.2\% (LLM-guided) versus 72.0\% (task-level cosine). The finer-grained decomposition into reusable subtask patterns provides more targeted guidance for completing individual tasks, particularly for Difficulty~3 tasks where subtask-level tips yield 58.7\% TGC versus 54.0\% for task-level (+4.7~pp).

\textbf{Retrieval strategy drives SGC.} LLM-guided selection dramatically improves scenario goal completion compared to cosine similarity at the same tip granularity: 64.3\% versus 57.1\% for subtask-level tips (+7.2~pp). This gap is consistent across difficulty levels, with Difficulty~1 showing the largest difference (89.5\% vs.\ 73.7\%, +15.8~pp). The LLM's ability to reason about task context, prioritize tip categories, and apply metadata filters produces more \emph{consistent} guidance across task variants within a scenario, reducing the sporadic failures that SGC penalizes.

\textbf{Interaction effect.} Interestingly, task-level tips with cosine similarity achieve higher SGC (62.5\%) than subtask-level tips with cosine similarity (57.1\%), despite lower TGC. Task-level tips encode holistic end-to-end strategies that promote uniform execution patterns across related task variants, while subtask-level tips---though more precise for individual task completion---may retrieve different subsets of subtask tips for different variants of the same scenario, introducing behavioral variance. LLM-guided selection compensates for this by reasoning about the overall task context and ensuring consistent tip selection across variants.

All configurations substantially outperform the baseline, confirming that the memory system provides genuine value regardless of the specific configuration chosen. The best configuration depends on the deployment objective: subtask-level tips with LLM-guided selection for the best overall performance, subtask-level tips with cosine similarity for the highest individual task accuracy at lower retrieval cost, or task-level tips with cosine similarity for a strong balance without LLM retrieval overhead.

\subsection{Source Partition Results (Train and Dev)}
\label{sec:source-partitions}

The train and dev partitions were used during tip generation: tips were extracted from agent trajectories on these tasks. Results on these partitions measure a distinct scenario from test-normal: \emph{what happens when the agent encounters the same or structurally identical tasks again, augmented with tips derived from its own prior executions?} This setting evaluates the memory system's ability to enable self-improvement on recurring tasks, complementing the generalization evaluation on test-normal.

Tables~\ref{tab:train-mem}--\ref{tab:dev-base} present results for subtask-level tips with LLM-guided selection on the source partitions.

\begin{table}[h]
\centering
\caption{Subtask Tips + LLM Selection: Train}
\label{tab:train-mem}
\begin{tabular}{l|c|c}
\hline
\textbf{Type} & \textbf{Task Goal} & \textbf{Scenario Goal} \\
\hline
Aggregate & 91.1 & 83.3 \\
Difficulty 1 & 94.4 & 83.3 \\
Difficulty 2 & 88.9 & 83.3 \\
Difficulty 3 & 88.9 & 83.3 \\
\hline
\end{tabular}
\end{table}

\begin{table}[h]
\centering
\caption{Baseline Agent (No Memory): Train}
\label{tab:train-base}
\begin{tabular}{l|c|c}
\hline
\textbf{Type} & \textbf{Task Goal} & \textbf{Scenario Goal} \\
\hline
Aggregate & 86.7 & 73.3 \\
Difficulty 1 & 100.0 & 100.0 \\
Difficulty 2 & 77.8 & 58.3 \\
Difficulty 3 & 77.8 & 50.0 \\
\hline
\end{tabular}
\end{table}

\begin{table}[h]
\centering
\caption{Subtask Tips + LLM Selection: Dev}
\label{tab:dev-mem}
\begin{tabular}{l|c|c}
\hline
\textbf{Type} & \textbf{Task Goal} & \textbf{Scenario Goal} \\
\hline
Aggregate & 89.5 & 73.7 \\
Difficulty 1 & 90.0 & 80.0 \\
Difficulty 2 & 87.5 & 62.5 \\
Difficulty 3 & 100.0 & 100.0 \\
\hline
\end{tabular}
\end{table}

\begin{table}[h]
\centering
\caption{Baseline Agent (No Memory): Dev}
\label{tab:dev-base}
\begin{tabular}{l|c|c}
\hline
\textbf{Type} & \textbf{Task Goal} & \textbf{Scenario Goal} \\
\hline
Aggregate & 77.2 & 47.4 \\
Difficulty 1 & 80.0 & 60.0 \\
Difficulty 2 & 70.8 & 25.0 \\
Difficulty 3 & 100.0 & 100.0 \\
\hline
\end{tabular}
\end{table}

As expected, improvements on the source partitions are larger than on test-normal: +4.4~pp TGC / +10.0~pp SGC on train, and +12.3~pp TGC / +26.3~pp SGC on dev. Tips are most contextually relevant when the agent encounters tasks structurally similar to those from which the tips were derived, so these larger gains are expected.

Two partition-specific patterns are worth noting. On train Difficulty~1 tasks where the baseline already achieves 100\%, the memory-enhanced agent scores slightly lower (94.4\% TGC, 83.3\% SGC), suggesting that for simple tasks where the agent already performs optimally, injecting additional tips can introduce minor interference. On dev, the Difficulty~3 baseline already achieves 100\% TGC and 100\% SGC, so the aggregate dev gains (+12.3~pp TGC, +26.3~pp SGC) are driven entirely by Difficulty~1 and~2 improvements. In both cases, the overall gains on the tasks that benefit from memory substantially outweigh any ceiling or interference effects.

\subsection{Cross-Configuration Summary}

Table~\ref{tab:summary} summarizes the aggregate improvements for subtask-level tips with LLM-guided selection across all three partitions.

\begin{table*}[h]
\centering
\caption{Summary of Aggregate Improvements: Subtask Tips + LLM Selection}
\label{tab:summary}
\begin{tabular}{l|c|c|c|c}
\hline
\textbf{Partition} & \textbf{Task Goal} & \textbf{Task Goal} & \textbf{Scenario Goal} & \textbf{Scenario Goal} \\
 & \textbf{(Baseline)} & \textbf{(+Memory)} & \textbf{(Baseline)} & \textbf{(+Memory)} \\
\hline
Test-Normal & 69.6 & 73.2 (+3.6) & 50.0 & 64.3 (+14.3) \\
Train & 86.7 & 91.1 (+4.4) & 73.3 & 83.3 (+10.0) \\
Dev & 77.2 & 89.5 (+12.3) & 47.4 & 73.7 (+26.3) \\
\hline
\end{tabular}
\end{table*}

Several observations emerge. First, the memory system improves performance on all three partitions, confirming that the benefits are not limited to tasks that generated the tips. The test-normal gains (+3.6 TGC, +14.3 SGC) demonstrate genuine generalization to unseen tasks. Second, the source partitions show larger TGC improvements, as expected—tips are most contextually relevant when the agent re-encounters tasks from which the tips were derived. Interestingly, the test-normal SGC gain (+14.3~pp) exceeds the train SGC gain (+10.0~pp), suggesting that the subtask-level decomposition and LLM-guided retrieval generalize particularly well for improving cross-variant consistency. Third, the SGC improvements consistently exceed the TGC improvements across all partitions, indicating that the memory system is particularly effective at improving consistency across task variants. Recovery tips and strategy tips encode prerequisite verification and error handling patterns that reduce behavioral variance, enabling the agent to reliably complete all variants rather than succeeding on some and failing on others.

\section{Related Work}

Our work sits at the intersection of agent memory systems, trajectory-based learning, and self-improving agents. We organize related work along three axes: memory architectures for LLM agents, systems that learn from execution trajectories, and approaches to agent self-improvement through experience.

\subsection{Memory Taxonomies and Architectures}

Two recent surveys provide comprehensive taxonomies of memory in LLM-based agents. Zhang et al.~\cite{zhang2024survey} organize the design space along three dimensions---memory \textit{sources} (agent-environment interactions, internal reasoning, user feedback), memory \textit{forms} (natural language, embeddings, databases, structured knowledge), and memory \textit{operations} (read, write, reflect, manage)---and identify key limitations of existing work: overly simplistic representations, unsophisticated operations for deciding what to remember or forget, and fragmented evaluation. Du et al.~\cite{du2025rethinking} take a complementary operations-centric view, defining six atomic memory operations: consolidation, updating, indexing, forgetting, retrieval, and compression. In their vocabulary, our tip extraction constitutes a form of \textit{consolidation} (converting raw trajectories into abstract tips), tip refinement is \textit{updating}, and selective retention is \textit{forgetting}. Both surveys note that most existing systems store raw or lightly processed text, lacking the structured abstraction and quality-aware curation that effective agent memory requires. Our framework directly addresses these identified gaps.

\subsection{Semantic Memory Systems}

The most widely deployed agent memory systems operate at the semantic level, storing factual knowledge extracted from interactions. Mem0~\cite{chhikara2025mem0} extracts and consolidates factual snippets---user preferences, entities, relationships---from conversations into a vector store, achieving strong latency and token efficiency for conversational personalization. A-MEM~\cite{xu2025amem} introduces a self-organizing memory architecture inspired by the Zettelkasten method, where each memory is stored as a structured note with contextual descriptions, keywords, and explicit links to related memories, creating an emergent knowledge network. While both systems are well-engineered for their purposes, they fundamentally store \textit{declarative} knowledge (what is known) rather than \textit{procedural} or \textit{experiential} knowledge (what to do and what was learned from doing it). They have no mechanism for analyzing execution trajectories, performing causal attribution of failures, or generating categorized behavioral guidance. Our framework addresses this gap by extracting structured, actionable tips from execution experience rather than conversational facts.

\subsection{Learning from Execution Trajectories}

A growing body of work addresses how agents can learn from their past execution traces, which is most directly related to our contribution.

\textbf{Workflow and procedure extraction.} Agent Workflow Memory (AWM)~\cite{wang2024awm} extracts reusable multi-step workflows from successful agent trajectories in web navigation, achieving 24.6\% and 51.1\% relative improvements on Mind2Web and WebArena respectively. AWM demonstrates a compelling ``snowball effect'' where simple workflows compose into more complex ones. However, AWM only learns from \textit{successful} trajectories---it has no mechanism for extracting lessons from failures, recoveries, or inefficient executions. Mem$^p$~\cite{fang2025memp} treats procedural memory as a first-class optimization object, systematically exploring strategies for building memory from trajectories, retrieving relevant procedures, and updating entries over time. While Mem$^p$ addresses the full memory lifecycle, it focuses on procedural instructions (``how to do X'') rather than the diagnostic behavioral insights (``what went wrong and why'') that our tip categories capture. AgentRR~\cite{feng2025agentrr} borrows the record-and-replay paradigm from software engineering, recording complete agent interaction traces and summarizing them into structured experiences for future replay. Like AWM, it primarily learns from successful executions.

\textbf{Reasoning and strategy extraction.} ReasoningBank~\cite{cai2025reasoningbank} is among the closest works to ours, distilling generalizable \textit{reasoning strategies} from an agent's self-judged successful and failed experiences. It shares our insight that agents should learn from both successes and failures. The key distinction is in abstraction level: ReasoningBank focuses on meta-cognitive reasoning strategies, while our tips focus on concrete behavioral guidance derived from specific execution patterns. The two approaches are complementary.

\textbf{Context engineering and self-improvement.} ACE (Agentic Context Engineering)~\cite{zhang2025ace} treats an agent's context as an evolving ``playbook'' that accumulates and refines strategies through a generate-reflect-curate cycle, achieving a 10.6 percentage point improvement on AppWorld. Our framework differs from ACE in several respects: we produce structured memory entries with typed categories (strategy, recovery, optimization), rich metadata, and selective retrieval rather than an evolving text document included in full; we perform explicit causal attribution tracing outcomes to specific decisions; and we maintain provenance tracking from tips to source trajectories.

\textbf{Experience replay with learned retrieval.} Memento~\cite{zhou2025memento} introduces a memory-augmented MDP formalization where a learned neural policy selects which stored trajectories to retrieve for a given task. However, Memento stores raw trajectories without abstracting them into transferable insights---the consolidation from trajectory to actionable lesson is left to the LLM's in-context reasoning.

\subsection{Empirical Foundations}

Xiong et al.~\cite{xiong2025memory} provide critical empirical grounding for trajectory-based memory systems, identifying the \textit{experience-following property} and two failure modes: \textit{error propagation} and \textit{misaligned experience replay}. They find that combining selective addition with selective deletion yields a 10\% absolute performance gain over naive memory growth. These findings directly motivate our structured approach: by extracting abstract tips with explicit applicability conditions rather than storing raw trajectories, and by categorizing tips with metadata for precise contextual matching, our framework mitigates both failure modes.

\section{Conclusions}

We presented a framework for automatically extracting actionable learnings from LLM-agent execution trajectories and storing them as structured memory tips that improve future agent performance. Our four-component pipeline---trajectory intelligence extraction, decision attribution analysis, contextual learning generation, and adaptive memory retrieval---captures the full spectrum of learning opportunities across failures, recoveries, inefficient successes, and clean successes. Evaluation on the AppWorld benchmark demonstrates consistent improvements, with up to 14.3 percentage point gains in scenario goal completion on held-out tasks, and particularly strong benefits on complex, multi-step tasks (28.5~pp SGC improvement, a 149\% relative increase). The framework naturally extends to multi-agent systems with cross-agent attribution and agent-role-aware guidance, which we leave to future work. We also plan to evaluate the framework with additional state-of-the-art and open-source models—such as Qwen~\cite{qwen2025qwen25} and GPT-OSS~\cite{gptoss2025}—to assess how tip quality and retrieval effectiveness vary across model families. The techniques described in this paper are being applied to IBM's Configurable Generalist Agent (CUGA)~\cite{cuga2025repo,marreed2025cuga} platform for building and deploying enterprise agentic systems, where trajectory-informed memory enables agents to continuously improve from operational experience.


\bibliographystyle{ACM-Reference-Format}
\bibliography{agentic-memory}

@article{zhang2024survey,
  title={A Survey on the Memory Mechanism of Large Language Model based Agents},
  author={Zhang, Zeyu and Bo, Xiaohe and Ma, Chen and Li, Rui and Chen, Xu and Dai, Quanyu and Zhu, Jieming and Dong, Zhenhua and Wen, Ji-Rong},
  journal={ACM Transactions on Information Systems (TOIS)},
  year={2025},
  note={arXiv:2404.13501},
  doi={10.1145/3748302}
}

@article{du2025rethinking,
  title={Rethinking Memory in {AI}: Taxonomy, Operations, Topics, and Future Directions},
  author={Du, Yiming and Huang, Wenyu and Zheng, Danna and Wang, Zhaowei and Montella, S{\'e}bastien and Lapata, Mirella and Wong, Kam-Fai and Pan, Jeff Z.},
  journal={arXiv preprint arXiv:2505.00675},
  year={2025}
}

@article{chhikara2025mem0,
  title={Mem0: Building Production-Ready {AI} Agents with Scalable Long-Term Memory},
  author={Chhikara, Prateek and Khant, Dev and Aryan, Saket and Singh, Taranjeet and Yadav, Deshraj},
  journal={arXiv preprint arXiv:2504.19413},
  year={2025}
}

@article{packer2023memgpt,
  title={{MemGPT}: Towards {LLMs} as Operating Systems},
  author={Packer, Charles and Wooders, Sarah and Lin, Kevin and Fang, Vivian and Patil, Shishir G. and Stoica, Ion and Gonzalez, Joseph E.},
  journal={arXiv preprint arXiv:2310.08560},
  year={2023}
}

@article{xu2025amem,
  title={{A-MEM}: Agentic Memory for {LLM} Agents},
  author={Xu, Wujiang and Liang, Zujie and Mei, Kai and Gao, Hang and Tan, Juntao and Zhang, Yongfeng},
  journal={arXiv preprint arXiv:2502.12110},
  year={2025}
}

@article{wang2024awm,
  title={Agent Workflow Memory},
  author={Wang, Zora Zhiruo and Mao, Jiayuan and Fried, Daniel and Neubig, Graham},
  journal={arXiv preprint arXiv:2409.07429},
  year={2024}
}

@article{feng2025agentrr,
  title={Get Experience from Practice: {LLM} Agents with Record \& Replay},
  author={Feng, Erhu and Zhou, Wenbo and Liu, Zibin and Chen, Le and Dong, Yunpeng and Zhang, Cheng and Zhao, Yisheng and Du, Dong and Hua, Zhichao and Xia, Yubin and Chen, Haibo},
  journal={arXiv preprint arXiv:2505.17716},
  year={2025}
}

@article{dechant2025risks,
  title={Episodic Memory in {AI} Agents Poses Risks That Should Be Studied and Mitigated},
  author={DeChant, Chad},
  journal={arXiv preprint arXiv:2501.11739},
  year={2025}
}

@article{xiong2025memory,
  title={How Memory Management Impacts {LLM} Agents: An Empirical Study of Experience-Following Behavior},
  author={Xiong, Zidi and Lin, Yuping and Xie, Wenya and He, Pengfei and Tang, Jiliang and Lakkaraju, Himabindu and Xiang, Zhen},
  journal={arXiv preprint arXiv:2505.16067},
  year={2025}
}

@article{fang2025memp,
  title={{Mem$^p$}: Exploring Agent Procedural Memory},
  author={Fang, Runnan and Liang, Yuan and Wang, Xiaobin and Wu, Jialong and Qiao, Shuofei and Xie, Pengjun and Huang, Fei and Chen, Huajun and Zhang, Ningyu},
  journal={arXiv preprint arXiv:2508.06433},
  year={2025}
}

@article{zhou2025memento,
  title={Memento: Fine-tuning {LLM} Agents without Fine-tuning {LLMs}},
  author={Zhou, Huichi and Chen, Yihang and Guo, Siyuan and Yan, Xue and Lee, Kin Hei and Wang, Zihan and Lee, Ka Yiu and Zhang, Guchun and Shao, Kun and Yang, Linyi and Wang, Jun},
  journal={arXiv preprint arXiv:2508.16153},
  year={2025}
}

@article{cai2025reasoningbank,
  title={{ReasoningBank}: Scaling Agent Self-Evolving with Reasoning Memory},
  author={Ouyang, Siru and Yan, Jun and Hsu, I-Hung and Chen, Yanfei and Jiang, Ke and Wang, Zifeng and Han, Rujun and Le, Long T. and Daruki, Samira and Tang, Xiangru and Tirumalashetty, Vishy and Lee, George and Rofouei, Mahsan and Lin, Hangfei and Han, Jiawei and Lee, Chen-Yu and Pfister, Tomas},
  journal={arXiv preprint arXiv:2509.25140},
  year={2025}
}

@article{zhang2025ace,
  title={Agentic Context Engineering: Evolving Contexts for Self-Improving Language Models},
  author={Zhang, Qizheng and Hu, Changran and Upasani, Shubhangi and Ma, Boyuan and Hong, Fenglu and Kamanuru, Vamsidhar and Rainton, Jay and Wu, Chen and Ji, Mengmeng and Li, Hanchen and Thakker, Urmish and Zou, James and Olukotun, Kunle},
  journal={arXiv preprint arXiv:2510.04618},
  year={2025}
}

@article{pink2025position,
  title={Position: Episodic Memory is the Missing Piece for Long-Term {LLM} Agents},
  author={Pink, Mathis and Wu, Qinyuan and Vo, Vy Ai and Turek, Javier and Mu, Jianing and Huth, Alexander and Toneva, Mariya},
  journal={arXiv preprint arXiv:2502.06975},
  year={2025}
}

@article{qwen2025qwen25,
  title={Qwen2.5 Technical Report},
  author={{Qwen Team}},
  journal={arXiv preprint arXiv:2412.15115},
  year={2025}
}

@misc{gptoss2025,
  title={{TODO: Add GPT-OSS reference}},
  author={{}},
  year={2025},
  note={Placeholder — please replace with the correct GPT-OSS citation}
}

@misc{cuga2025repo,
  title={{CUGA}: Configurable Generalist Agent},
  author={{IBM}},
  year={2025},
  howpublished={\url{https://github.com/cuga-project/cuga-agent}}
}

@article{marreed2025cuga,
  title={Towards Enterprise-Ready Computer Using Generalist Agent},
  author={Marreed, Sami and Oved, Alon and Yaeli, Avi and Shlomov, Segev and Levy, Ido and Akrabi, Offer and Sela, Aviad and Adi, Asaf and Mashkif, Nir},
  journal={arXiv preprint arXiv:2503.01861},
  year={2025}
}



\end{document}